\pdfoutput=1

\documentclass[11pt]{article}

\usepackage[final]{acl}

\usepackage{times}
\usepackage{latexsym}

\usepackage[T1]{fontenc}

\usepackage[utf8]{inputenc}

\usepackage{microtype}

\usepackage{inconsolata}

\usepackage{graphicx}
\usepackage{subfigure}
\usepackage{bbm}
\usepackage{booktabs}
\usepackage{amsmath}
\usepackage{amssymb}
\usepackage{mathtools}
\usepackage{amsthm}
\usepackage{xspace}
\usepackage{enumitem}
\usepackage{hyperref}
\makeatletter
\def\thanks#1{\protected@xdef\@thanks{\@thanks
        \protect\footnotetext{#1}}}
\makeatother
\newcommand{\chen}[1]{\textcolor{cyan}{[Chen: #1]}}
\newcommand{\kevin}[1]{\textcolor{purple}{[Kevin: #1]}}
\newcommand{\haotian}[1]{\textcolor{red}{[Haotian: #1]}}

\newcommand{\eat}[1]{}
\title{EPO: Hierarchical LLM Agents with Environment Preference Optimization}

\author{Qi Zhao\textsuperscript{*}, Haotian Fu\textsuperscript{*}\thanks{*: Equal contribution. Code and dataset can be found at~\url{https://github.com/kevinz8866/EPO}.}, Chen Sun, George Konidaris \\ \\
  Brown University \\}

\begin{document}
\maketitle
\begin{abstract}

Long-horizon decision-making tasks present significant challenges for LLM-based agents due to the need for extensive planning over multiple steps. In this paper, we propose a hierarchical framework that decomposes complex tasks into manageable subgoals, utilizing separate LLMs for subgoal prediction and low-level action generation.  To address the challenge of creating training signals for unannotated datasets, we develop a reward model that leverages multimodal environment feedback to automatically generate reward signals. We introduce Environment Preference Optimization (EPO), a novel method that generates preference signals from the environment's feedback and uses them to train LLM-based agents. Extensive experiments on ALFRED demonstrate the state-of-the-art performance of our framework, achieving first place on the ALFRED public leaderboard and showcasing its potential to improve long-horizon decision-making in diverse environments.
\end{abstract}
\section{Introduction}
Long-horizon decision-making/planning remains a formidable challenge for Large Language Model(LLM)-based agents~\citep{DBLP:conf/nips/ValmeekamMSK23, DBLP:journals/corr/abs-2304-11477, DBLP:conf/aaai/SilverDSTK024}. These tasks require extensive planning over multiple steps, maintaining coherence and goal orientation, which is difficult for LLMs that are typically designed for more immediate and localized predictions. Moreover, a key issue of finetuning LLMs for embodied agents is the need of large scale labeled data \citep{reed2022generalist}. The same issue is reflected in researchers' effort in building reward models from vision foundation models as we might need to obtain ``internet-scale'' data of task demonstrations \citep{fan2022minedojo}.

To tackle the first challenge, a straightforward way is to first let the LLM decompose the long-horizon task into shorter horizon subtasks, and then use different LLMs as the policies at different levels, i.e., use one LLM-based policy to generate subgoals, and use another LLM generate low-level actions given the subgoals, both of which require significantly fewer planning steps. This decomposition facilitates more effective planning and execution by leveraging the predictive power of LLMs at both the subgoal and action levels. 

\eat{\chen{From Figure 1, it seems that everything is done in language space and perception signals are expressed as ``symbolic representation''. This is cool that it works, but readers might have questions whether this is a realistic setup, or more an artifact from ALFRED dataset.}\haotian{That is true. Do you have any suggestions for better expressing this?}}

However, the problem of how to efficiently train these LLM-based agents remains. In this paper, we consider the setting where only part of the dataset are annotated with ground-truth actions and subgoals, and we need to find a way to create training signals for the unannotated dataset. The common training signals for decision-making agents are based on the rewards received during interactions with the environment~\citep{DBLP:books/lib/SuttonB98}. But the manual design of reward functions is both time-consuming and prone to inaccuracies, which hinders the scalability and adaptability of LLM-based agents in dynamic and diverse environments.  Consequently, there is a growing need for methods that can automatically generate reward signals from the environment, thus bypassing the complexities associated with human-engineered rewards. This motivation drives us to explore reward modeling approaches that can leverage multimodal feedback from the environment, such as visual and interaction data, to guide the learning process of LLM-based agents by leveraging the public pretrained foundation models.

On the other hand, recent advancements in preference optimization techniques, such as Direct Preference Optimization (DPO)~\citep{rafailov2023direct}, have shown that LLMs can be effectively trained using preference-based signals rather than explicit reward functions. DPO leverages the inherent capabilities of LLMs to model preferences between different outputs, facilitating a more intuitive and flexible training paradigm. This insight inspires us to develop a novel method that combines the strengths of preference optimization with automatic reward modeling to enhance the performance of LLM-based agents in long-horizon decision-making tasks.

In this paper, we propose a hierarchical LLMs-based framework for long-horizon decision making problems. Our agent decomposes complex tasks into manageable subtasks by training two LLMs to predict the subgoal decomposition and low-level actions respectively. To retrieve enough training signals from the unannotated dataset, we propose a LLM-based reward model that is able to integrate the multimodal environment feedback information and automatically generate reward signals for the unannotated dataset. Then, we introduce Environment Preference Optimization (EPO), a method that generates preference signals automatically from the environment's feedback. EPO ranks the proposed actions and subgoals based on the estimated rewards and constructs a preference dataset that guides the training of LLM-based agents. This approach leverages both annotated and unannotated datasets, significantly expanding the training data available for improving agent performance. 


To validate our framework design, we conduct extensive experiments on ALFRED~\citep{shridhar2020alfred}, a popular household simulation environment for embodied agents. Our method achieves the state-of-the-art performance on ALFRED. We also find that unified environment feedback significantly help decision-making agents in both subgoal decomposition level and environment interaction level. Moreover, in the setup where there exists a large dataset of task specifications but only a small annotated task and demonstrations, our framework allows agent to benefit from the unannotated new tasks while significantly outperforming supervised training, indicating the potential of our framework. 

To sum up, we make the following contributions: 

\begin{enumerate}
    \item We propose a hierarchical LLMs-based framework for long-horizon decision-making problems, where both levels of LLMs can be jointly trained with preference signals generated from a LLM-based reward model.
    
    \item We propose Environment Preference Optimization (EPO), a method that first learns to automatically generate preference signals for an unannotated dataset from multimodal environment feedbacks by learning a reward model, and then use them to train/finetune the hierarchical LLMs-based agents.
    
    \item We demonstrate the effectiveness of our framework through extensive experiments and achieved state-of-the-art performance on ALFRED (we reached {\bf the first place on the ALFRED public leaderboard}\footnote{\url{https://leaderboard.allenai.org/alfred/submissions/public}. EPO has been top of the leaderboard as of the release date of this paper.}).
\end{enumerate}

\eat{\chen{The intro reads nice, and my main comment is that you might want to justify the baseline framework (LLM-based agents) a bit so your contributions can be better positioned and appreciated. Three example questions: 1. hierarchical decision making is widely used, are we the first to propose hierarchical llm-based agent? 2. Among the three prior work at the beginning of intro, what are their commonalities to your baseline framework, what are their problems that need to be addressed by you? 3. Why is it desirable to have LLM produce low-level policies? (argubly, the next action and object are not ``low-level'' enough)}\kevin{for the third question, I guess it is because a LLM-based low-level policy could handle diverse subgoals expressed in language. Idealy we would want this subgoal space to be very large.}}
\section{Related Work}

\textbf{Foundational Models for Embodied Agents.} A number of recent works have explored foundational models for embodied agents~\citep{DBLP:conf/icml/DriessXSLCIWTVY23, DBLP:conf/corl/StoneXLGLVWKZXF23, DBLP:conf/rss/BrohanBCCDFGHHH23, DBLP:conf/corl/ZitkovichYXXXXW23}. Our work is inspired by many 
previous language grounding agents work \citep{singh2023progprompt, ahn2022can, huang2023inner}
on robotics. These studies work on grounding natural language prompt or robotic actions with symbolically represented visual or interaction information. Similarly effort in grounding language to visual information for embodied agents have been done in \citep{song2023llm}. Among works in simulation, \citet{pashevich2021episodic} present the end-to-end approach for decision-making agents, which directly predicts the agent's next action from task specification and visual input without subgoal alignment and map-based navigation. \citet{min2021film} introduce a hierarchical approach, which has dominated due to their superior performance. \citet{DBLP:journals/corr/abs-2402-16354} leverages LLM to help learning skills from demonstrations. Our hierarchical LLMs framework is also inspired by many prior hierarchical RL works~\citep{DBLP:conf/nips/NachumGLL18, DBLP:conf/iclr/LevyKPS19, DBLP:conf/icml/FuYTL023}.

\textbf{Reward Modeling with Foundational Models.} Foundation models with their capability in encoding generic representations of a modality have motivated researchers to use them to generate reward signals in order to bypass human reward engineering. Among these efforts, \citet{sontakke2023roboclip, escontrela2023video, chen2021dvd, fan2022minedojo, pmlr-v162-mahmoudieh22a} use vision foundation models to estimate the reward by aligning visual features with desired actions or state transitions. However, these approaches often require large scale data. In contrast, we are interested in using pretrained LLMs to generate reward signals \citep{kwon2023reward} from all symbolically represented environment feedback. Within this scope, \citet{song2023self, yu2023language, ma2023eureka, huang2023voxposer, DBLP:journals/corr/abs-2311-01455} use language models to generate rewards to help robot learn skills based on the symbolic states. For embodied agents, ELLM \citep{du2023guiding} propose a framework to use LLMs to guide agents' exploration and generate reward based on the task goals in 2D games and robotic simulators. Compared to existing works, we fill in the blank by proposing a generic framework that use LLMs to synthesize reward from multimodal environment feedback. 

\textbf{Preference-Based Learning for Language Models.} Aligning language models to human preference \citep{ouyang2022training} has greatly improved language models to follow human instructions. Recent development such as Direct Preference Optimization \citep{rafailov2023direct}, self-rewarding language models \cite{yuan2024self} in preference alignment allows the language model to directly learn the preference relation and also learn from its own synthesized data. Inspired by these work, we extend the definition of ``preference'' into the alignment between environment feedback and agent actions with respect to the task specification. We leverage the algorithmic advantage demonstrated in DPO and the idea of self data synthesis~\citep{lee2023rlaif} to train LLM-based embodied agents to ground language to environment feedback.
\section{Method}

\begin{figure*}[t]
\begin{center}
\centerline{\includegraphics[width=\textwidth]{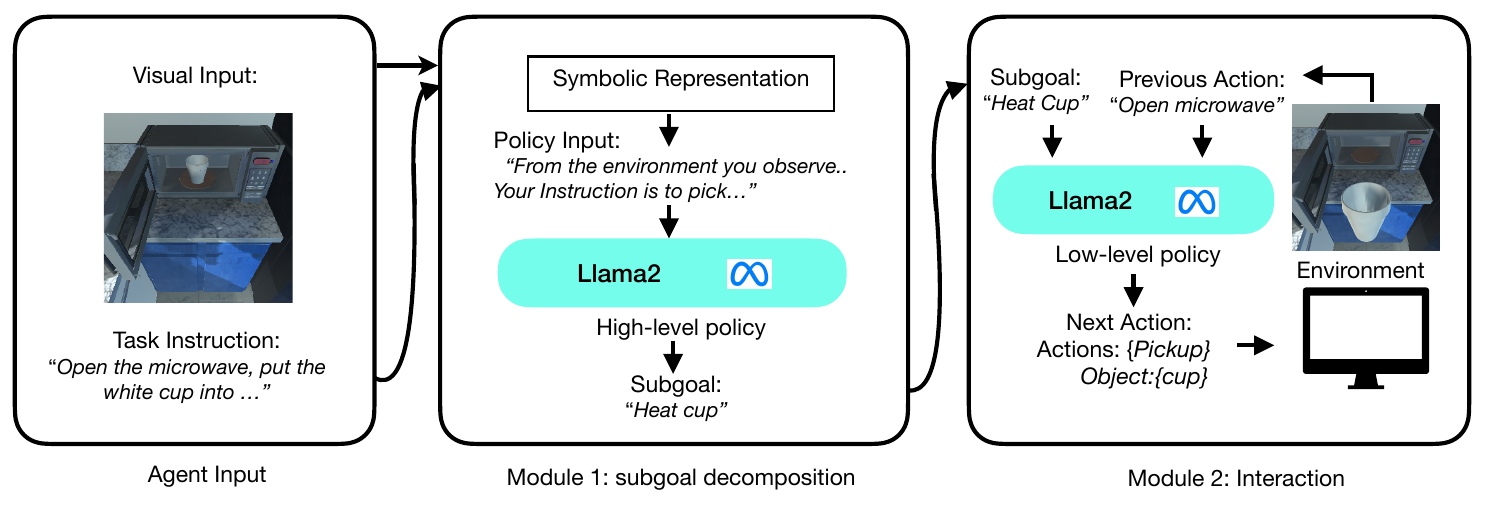}}
\vspace{-0.1in}
\caption{An illustration of the hierarchical framework. Our agent first outputs the subgoals from human instructions and visual inputs using its high-level subgoal decomposition module. Then the interaction module predicts low-level actions autoregressively to complete the given subgoals.}
\label{main_vis}
\end{center}
\vspace{-0.4in}
\end{figure*}

We first describe the problem setup in~\ref{sec:setup} and then introduce our hierarchical LLMs-based decision-making agent in~\ref{sec:hierarchy}. Then we present our approaches for generating reward signals from multimodal environment feedback in~\ref{sec:rew_model}. Lastly, we explain how we train the hierarchical agents with Environment Preference Optimization in~\ref{sec:epo}.

\subsection{Problem Setup}
\label{sec:setup}
In this paper, we consider the decision-making agents that take in human language instructions $G$ as well as visual observations $o$ from environment $E$, and generate a sequence of actions $a$ to interact with the environment, aiming to achieve the goal described by $G$. Low-level action $a$ is parameterized by an action type $l$ and optionally a target object $k_l$, in the form of natural language, e.g. Pickup (apple), Moveforward (None). 
We consider the setting of learning from demonstrations and we have access to the environment $E$ that each task is associated with, where the reward function is not provided if we let the agent interact. We assume the agent is given a {\bf partially-annotated} dataset--- a certain portion of the dataset are unannotated. Each trajectory from the fully annotated part of the dataset  consists of $\{G, E, g_1, a_1, g_2, a_2,\cdots\}$, where $g_t$ denotes the assigned subgoal for current timestep $t$. $g$ from the dataset is also described by language. The unannotated part of the dataset consists of task goals and environments (no reward function) without the ground-truth low-level actions and subgoals $\{G_1, E_1, G_2, E_2, \cdots\}$. The performance of our agent is measured with task success rate, which is the percentage of test tasks completed given a set of human task instructions.
\eat{\chen{Reader might find our definition of ``unannoatated'' dataset a bit confusing since the subgoals are given.}\kevin{We actually don't have the g1,g2 for the unannotated dataset @haotian}\haotian{Changed}}

\subsection{Hierarchical LLMs-based Agent}
\label{sec:hierarchy}
LLMs are known for struggling with long-horizon planning tasks. A natural way to alleviate this issue is by decomposing the tasks into shorter-horizon subtasks. We show our hierarchical LLMs-based agent framework in Figure~\ref{main_vis}. We finetune \eat{\chen{from scratch? if not, how would pre-training help embodied agent?} \haotian{We finetune pretrained LLMs. Both the input and output of our model are language-based so pretrained LLMs can help.}}pretrained LLMs to output predictions for subgoals given the general task goal, and finetune another LLM to output predictions for low-level actions given the subgoals. Specifically, we parameterize each subgoal with a high-level action type $h$ and a target object/position $k_h$, both in the form of language, similar to what we set for the low-level actions. Note that the subgoal may look same to some low level actions, e.g. ``pickup potato''. However, the ``pickup'' low level action can be executed only when the agent is at a place near the potato and facing towards it, while the subgoal ``pickup potato'' needs to be executed from anywhere and may require many low-level actions for navigation. Given the task instruction $G$ (e.g. Wash the apple on the counter) and the original subgoal described by natural language (e.g., Find the apple), the high-level decomposition module (parameterized by an LLM) $\pi_{h}$ outputs the decomposed subgoals $\{h, k_h\}= \pi_{h}(G, g)$, e.g. ``Heat Cup''. 

We find this subgoal decomposition design especially beneficial for training embodied agents that directly use LLMs as their policies since: 1. Subgoals with a fixed form instead of the free-form language from the dataset enable us to better infer the preference signals between two possible responses (see Section~\ref{sec:epo}). 2. It functions as a translation of the original subgoal instructions described in natural language. E.g., we find that in practice, in ALFRED, one of the subgoal instructions given be the dataset is ``Then, pick up the dog from the desk''. However, there's no dog in the room and the ``dog'' in the instruction actually refers to the statue that looks like a dog. Thus our subgoal decomposition outputs two subgoals  ``Moveto desk'' and ``Pickup statue'', which correct the mistake in the dataset and also make the subgoals more concise for the low-level policy (LLM) to infer the grounding actions. 

For the low-level interaction module, the agent is given the subgoal decomposition output from the high-level module, and autoregressively outputs a sequence of low-level actions $a$, each of them parameterized by an action type $l$ and an target object/position $k_l$ to interact with the environment. At timestep $t$ for a given transformed subgoal $\{h, k_h\}$ and sequence of past actions in completing this subgoal $a_{\text{past}} = \{a_{0}, \cdots ,a_{t-1}\}$, the low-level interaction module $\pi_{l}$ predicts the next low-level action $a_t = \pi_{l}(h, k_h, a_{\text{past}})$ to reach the given subgoal, all in the form of natural language. The low-level agent will output a <stop> token if it thinks the subgoal is fulfilled - the language-model-based policy outputs a sequence of actions and we switch to the next subgoal once these actions are all executed. One can expect the agent complete the given task if its subgoal decomposition module can predict the subgoal sequence correctly and for each subgoal, the skill module can output the correct low-level action sequences.

\begin{figure*}[t]
\vspace{-0.3in}
\begin{center}
\centerline{\includegraphics[width=\textwidth]{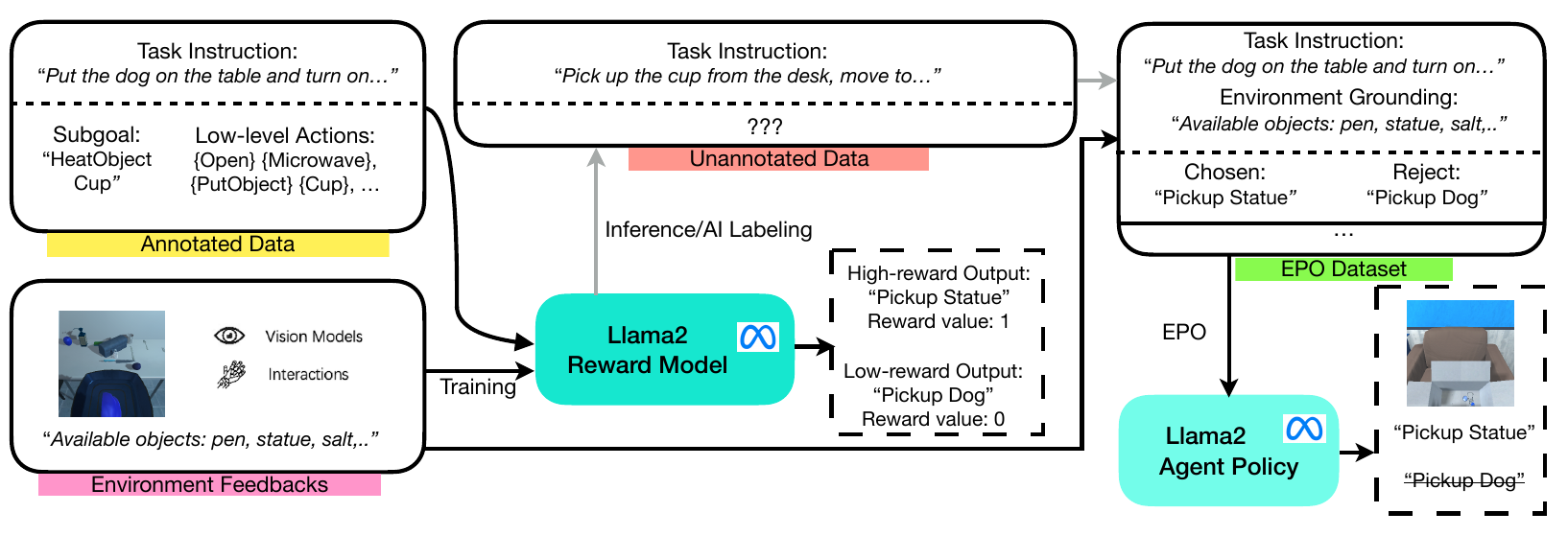}}
\caption{An illustration of our pipeline to train reward model for grounding environment feedback with human instructions. We supervisedly train the reward model given the annotated data. Then we use the reward model to label unannotated data to obtain the preference relations. Then we form the EPO datasets and optimize our agent policies using the proposed EPO algorithm.}
\label{main_vis2}
\end{center}
\vspace{-0.4in}
\end{figure*}

\subsection{Reward modeling from Environment Feedback}
\label{sec:rew_model}
One of the key motivations of this paper is to bypass the complex human-based reward engineering and learn to automatically generate feedback signals for a diverse set of unannotated tasks that can help train the LLM-based agent. To this end, we propose an approach to learn a reward model that is able to generate feedback signals from the multimodal observations of the environment. We show the proposed Reward Modeling and EPO training framework in Figure~\ref{main_vis2}. 

\textbf{Environment Feedback.} We consider two types of environment feedback that an embodied agent can typically receive. The first one is visual positional feedback, i.e., each timestep the agent will receive a visual observation (image) describing the current environment, and we apply pretrained vision models to retrieve visual positional feedback $V$ in the form of labels or natural language. For example, given an observed frame in a house, an object detection model of our agent will output a list of objects detected from its label space or a textual description such as "a computer on top of a desk". The second type of feedback is interaction feedback. If the agent attempt to interact with the detected objects using low-level actions like ``pick up'' or ``close'', it will receive a interaction feedback $I$ in the form of boolean values or natural language. For example, our agent could attempt to ``Pick up Cup'', then it will receive a boolean value indicating if its action succeeded.

\textbf{Reward Modeling.} In order to unify the feedback information, we symbolically represent them all in language if they are in the form of labels. We denote the language represented feedback information as $F$. Our reward model $R_{\rho}$ takes in the feedback information $F$, task specific input $T$ and predicted subgoal/action $P$ from the LLM, and outputs a reward value which describes the alignment score of the proposed output with respect to the task input, given newly observed environment feedback. Here, feedback information $F$ can be visual feedback $V$, interaction feedback $I$, or both. Task specific input $T$ can be the input of the high-level decomposition module $\{G, g\}$ or that of the low-level interaction module $\{h,k_h, a_{\text{past}}\}$. Predicted output $P$ can be the output of the subgoal decomposition module $\{h, k_h\}$ or that of the interaction module $a$.
\begin{equation}
\small
    \hat{r} = R_{\rho}(F, T, P)
\end{equation}
To train this reward model, we construct positive pairs based on whether the proposed output is correct with respect to the task input and assign them with high rewards. Similarly we construct negative pairs with incorrect proposed output and low rewards. For instance, if the visual positional feedback we get from the environment after symbolic representation $F$ is ``there exists a cup and an apple on the counter'' and our task instruction $T$ is ``pick up the red apple on the left side of the cup'', we will construct the positive pair using the correct label, so our proposed answer $P$ is ``Pick up object apple''. When constructing the negative pair, we have the same $F$ and $T$, but the proposed answer is randomly chosen from possible outputs, it can be ``Pick up object cup''. In this way, we construct a synthetic dataset that maps the environment feedback, task specifications, and proposed answers to reward values. Then we train the reward model using the cross-entropy loss.

\subsection{Environment Preference Optimization}
\label{sec:epo}
 With the trained reward model, we can leverage the unannotated dataset by evaluating our agent's proposed subgoals or low-actions according to the given environment feedback and task specification. We first pretrain the hierarchical LLM modules on the annotated dataset. Then on the unannotated dataset, we use our reward model to evaluate the LLM modules' outputs and rank them according to the estimated reward. After that, we will have a ranking of the outputs $ (p_1, p_2, \ldots, p_n)$, where $p_1$ denotes the output that is given the highest reward: $\hat{r}_{p_1} = \max{\hat{r}_{p_i}}$. It holds that \( \hat{r}_{p_i} > \hat{r}_{p_j} \) if \( i < j \).

\noindent From the response ranking, we can construct a preference dataset $\mathcal{D} = \{ (F_1, T_1, p_{w1}, p_{l1}), (F_2, T_2, p_{w2}, p_{l2}), \ldots \}$, where $p_{w1}$ is the proposed output that is more likely correct, $p_{l1}$ is the less likely one. Given that the environment feedback and our reward model labeling might not be perfect, especially under the circumstance of insufficient labeled data, we propose Environment Preference Optimization (EPO) which combines DPO~\citep{rafailov2023direct} training with an token-level alignment loss. We provide additional token-level constraint while preserving the learning of preference relations. The training objective is as below:
\vspace{-0.00in}
\begin{equation}
\label{eqn:epo}
\small
    \mathcal{L}_{\text{EPO}}(\theta) = \mathbb{E}_{(T, p_{w}, p_{l}) \sim \mathcal{D}} \left[- p_{w} \log(\pi_{\theta}(\hat{p}\mid T)) + \mathcal{L}_{D}  \right]
\end{equation} 
, where
\begin{equation}
\small
\begin{aligned}
&\mathcal{L}_{D} = -\mathbb{E}_{(T, p_w, p_l) \sim \mathcal{D}} \Big[ \log \sigma \Big( \beta \log \frac{\pi_\theta(p_w \mid T)}{\pi_{\text{sup}}(p_w \mid T)} \\ &- \beta \log \frac{\pi_\theta(p_l \mid T)}{\pi_{\text{sup}}(p_l \mid T)} \Big)\Big].
\end{aligned}
\end{equation}
$\pi_{\theta}$ denotes the LLM we are trying to optimize and it can be either the subgoal decomposition module $\pi_h$ or the low-level interaction module $\pi_l$. $\sigma$ denotes the logistic function. $\beta$ is the hyperparameter for scaling. We use $\pi_{\text{sup}}$ to denote the LLM learned from the annotated dataset and denote the logits of our model output tokens as $\hat{p}$. The training objective of $\mathcal{L}_{D}$ is to maximize the log probability difference between the chosen response and the rejected response, which is calculated based on all tokens. Note that DPO does not force the model to align with the chosen output, instead it encourages the model to maximize the reward difference between chosen outputs and rejected outputs---it does ``soft-alignment''. However, in our case we still want our model to "hard-align" to the labels with the highest reward since they are mostly likely to be the correct label. Furthermore, we want to reduce the algorithmic instability rises in ``soft-alignment'', which could hamper LLMs to follow certain desired output format. For example in practice, we want our high-level subgoal decomposition policy to output both of the subgoal parameters $h$ and $k_h$. With the alignment loss (first term in Eqn~\ref{eqn:epo}), we guide the optimization process to reduce the algorithmic instability rises especially when we train with a large amount of unlabeled data. In this way, we let the model learn the preference relation between answers but also align towards the most correct outputs with parameters in given format since it does the reward modeling and the token level optimization at the same time. Note that in practice, we apply EPO to both high- and low-level policies’ training process. 
\section{Experimental Details}


\subsection{Environment} We conduct experiments on ALFRED \citep{shridhar2020alfred}, a popular household simulation environment based on AI2-THOR~\citep{DBLP:journals/corr/abs-1712-05474} for embodied agents. It consists of 120 indoor simulations of different room types. The official expert demonstration dataset consist of 8055 task demonstration annotated with 25,743 natural language instructions in English. The entire dataset is split into 21023 instructions in training set, 820 in seen validation set whose environment scenes are shared with those in the training set, 821 in unseen validation whose environment scenes are not available in training set. Only the task instructions in training and validation set are paired with the subgoal and low-level action annotations. Subgoals and actions annotations are in the form of structured natural language. In this environment, our agent receives egocentric visual observation in RGB, and render low-level actions to interact with the environment. The low-level action space consists of 12 discrete action types and 82 discrete object types. 

\subsection{Implementation details}

We use pretrained RCNN as the object detection model and Mask-RCNN as the segmentation model~\citep{he2017mask}. For representing visual information, we also want to study how visual detail information (e.g. image captions) could contribute as a form of environment feedback. Therefore, we use BLIP-2~\citep{li2023blip} as our image captioning model and we apply it at the view-points where we can interact with the objects.

For both levels of our agent modules, and reward models, we use Llama2-7B~\citep{touvron2023llama} as the large language model backbone and use LoRA~\citep{DBLP:conf/iclr/HuSWALWWC22} to efficiently finetune the language models. 


\textbf{Agent Learning.} In order to validate the effectiveness of our framework in learning from unannotated dataset, we split the annotated trained dataset into a labeled dataset for which we have access to the annotated labels and a unlabeled dataset for which we have only access to the task specifications without labels, to mimic the real world scenario where we have only limited annotated expert demonstrations but can access to many new task specifications. On the unlabeled dataset, we use our reward model trained on the labeled dataset to inference reward for each possible outputs. Then we form the environment preference dataset based on the rewards of the outputs. 
More details about our experimental setting can be found in the appendix.

\section{Results}
\begin{table*}
\centering
\small
\scalebox{1.0}{
\begin{tabular}{c cccccccc}
\toprule
\multicolumn{1}{c}{} & \multicolumn{2}{c}{Success Rate} & \multicolumn{2}{c}{GC} & \multicolumn{2}{c}{PLWSR} & \multicolumn{2}{c}{PLWGC} \\
\midrule
\ Model & Unseen & Seen & Unseen & Seen &  Unseen & Seen & Unseen & Seen \\ 
\midrule
HLSM~\cite{blukis2022persistent} & 0.2027 & 0.2994 & 0.3031 & 0.4121 & 0.0555 & 0.0874 & 0.0999 & 0.1458 \\
FILM~\cite{min2021film} & 0.2780 & 0.2883 & 0.3852 & 0.3955 & 0.1132 & 0.1127 & 0.1513 & 0.1559\\
EPA~\cite{liu2022planning} & 0.3607	& 0.3996 & 0.3954 & 0.4414 & 0.0292 & 0.0256  & 0.0391 & 0.0347\\
Prompter~\cite{inoue2022prompter} & 0.4572 & 0.5323 & 0.5876 & 0.6343 & 0.2076 & 0.2581  & 0.2622 & 0.3072\\
CAPEAM~\cite{kim2023context} & 0.5036 & 0.5258 & 0.6140 & 0.6098 & 0.2159 & 0.2309  & 0.2531 & 0.2710\\
\textbf{EPO (ours)} & \textbf{0.6235} &\textbf{0.6479} & \textbf{0.6752} & \textbf{0.7230} & \textbf{0.5199} & \textbf{0.5692} & \textbf{0.6415} & \textbf{0.6620}\\
\bottomrule
\end{tabular}
}
\caption{Comparison with SOTA methods on ALFRED test set. GC stands for ``goal-conditioned''. PLW stands for ``path length weighted''. We get the data of the baselines from ALFRED's public leaderboard.}
\label{tab:sota}
\end{table*}

In this section, we first compare the overall performance of our framework with the state-of-the-art methods on ALFRED public leaderboard and then modularly study the components of our framework. We obtain all the results following the standard setting in ALFRED where we first let the agent learn from the given dataset offline, and then test the the {\bf online rollout performance} of the learned policies (modules) on the given set of new test tasks.

\subsection{Comparison with SOTA on ALFRED}
To demonstrate the effectiveness of our framework, we compare the performance of the proposed algorithm to existing works on ALFRED public leaderboard on the hold out test set. Here we use the best setup for all our module. That means we use the subgoal decomposition module and interaction module both trained on environment feedback with reward modeling and EPO. In Table \ref{tab:sota}, our method significantly outperforms previous work over 12.0\% on unseen test tasks while achieving SOTA performance on both unseen and seen scenarios in all metrics, indicating the effectiveness of our approach. Moreover, our method achieves significant superior performance on path length weighted (PLW) metrics, which indicates the efficiency of our method in completing the tasks in fewer steps. It is worth mentioning that our approach does not use semantic voxel map \citep{shridhar2020alfworld}, which requires the access of environment meta data. Our approach uses agent exploration (Appendix~\ref{app:imp}) to obtain object location information which generalizes better to real world scenarios without the meta information defined in simulators. 

\subsection{How well does EPO learn from unannotated data?}
\textbf{Environment preference optimization enhances the agent's performance via training on the unannotated data.} We compare to Supervised Fine-Tuning (SFT), where we directly prepend the environment feedback to task information and train only use the annotated dataset. To study whether our proposed framework can further improve itself through learning from unannotated dataset, we consider three data split. First, full/No-Split means we use the entire annotated ALFRED dataset. Second, 90/10 means we use 90\% of the demonstration with their annotations and 10\% of the demonstration without annotation. Lastly, 10/90 refers to the split where only 10\% of the data we use is annotated and 90\% is unannotated. We can see that in all three setups, our method based on environment preference optimization outperforms supervised fine-tuning. As we increase the amount of unannotated data, one can observe that our framework start to show more significant superior performance than supervised fine-tuning. This trend of our proposed EPO performing better when there exists more unannotated data indicates that the data efficiency and potential of EPO in real application scenarios, as data efficiency is one of the most important problems for learning from demonstrations in practice. 


\begin{table}[h]
\small
\centering
\scalebox{1.0}{
\begin{tabular}{cccc}
\toprule
\ Learning & Data Split & Unseen & Seen \\ 
\midrule
SFT & full & 0.5383 & 0.4939 \\
EPO & full & \textbf{0.5481} & \textbf{0.5024} \\
\midrule
SFT & 90/10 & 0.5286 & 0.4841 \\
EPO & 90/10 & \textbf{0.5445} & \textbf{0.4988} \\
\midrule
SFT & 10/90 & 0.4689 & 0.4305 \\
EPO & 10/90 & \textbf{0.5091} & \textbf{0.4668} \\
\bottomrule
\end{tabular}
}
\caption{Comparing different learning paradigms on validation dataset. }
\label{tab:algorithm}
\vspace{-0.2in}
\end{table}

\begin{figure*}[t]
\vspace{-0.5in}
\begin{center}
\centerline{\includegraphics[width=\textwidth]{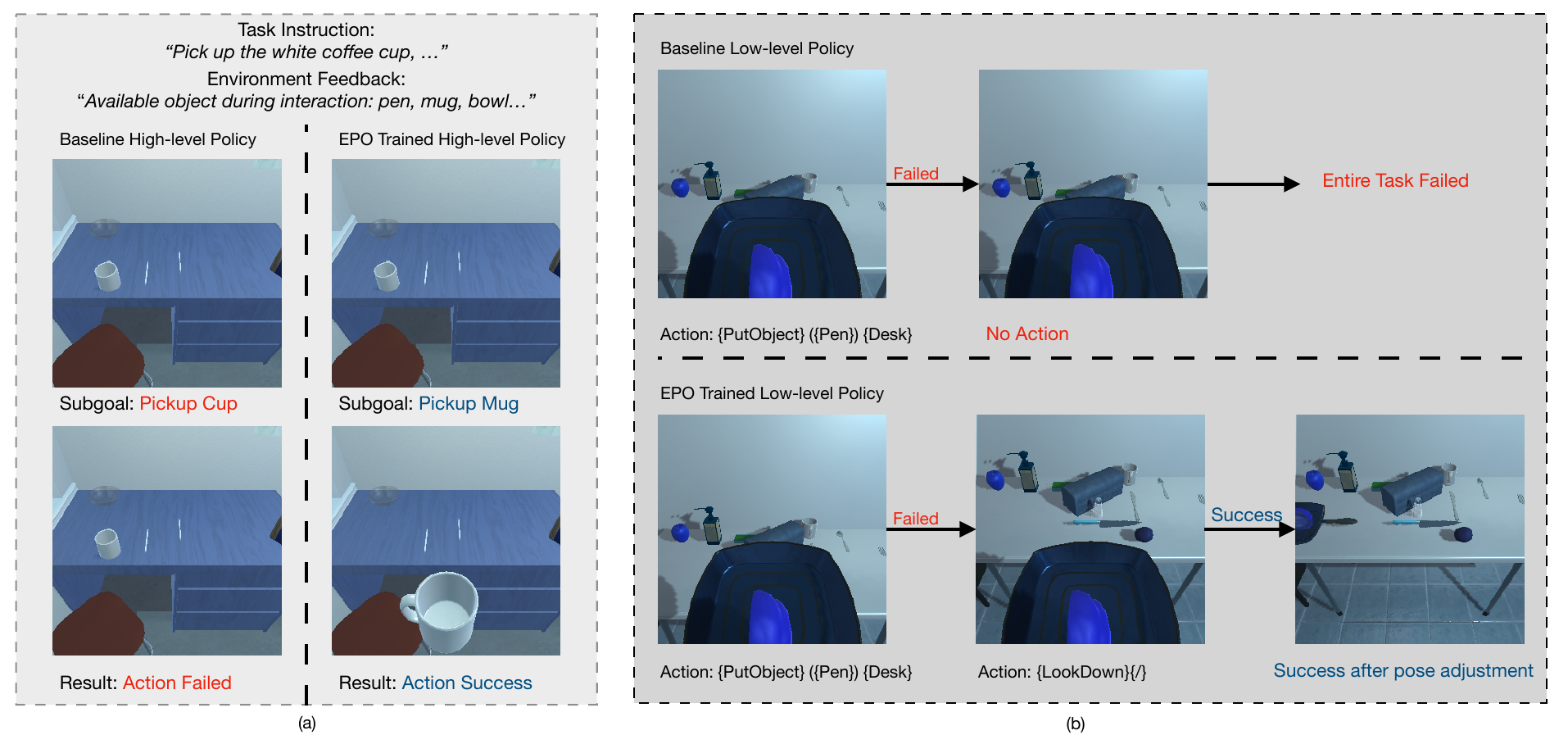}}
\vspace{-0.1in}
\caption{An visual illustration of how EPO improved both high-level subgoal decomposition policy and the low-level interaction policy. In the left figure, we present the difference between a baseline high-level policy and a EPO trained counterpart. We observe that the latter one can correctly figure out the subgoal. In the right figure, we present the difference between a baseline low-level policy and a EPO trained counterpart. We observe that the latter one can conduct post adjustment to successfully execute the actions.}
\label{main_vis3}
\end{center}
\vspace{-0.2in}
\end{figure*}
\subsection{How well do different environment feedbacks help decision making?}

\textbf{Reward modeling can help improve low-level interaction module.} Previous work on ALFRED \citep{min2021film} makes the hypothesis that, ALFRED's low-level action dynamics to accomplish the interaction subgoals are quite deterministic and can potentially be handled with a deterministic program. We consider the comparison between our learning-based interaction module (LLM) against the hard-coded deterministic program. Here we use the same subgoal decomposition policy which is supervised fine-tuned with the environment feedback and only change the interaction module for a fair comparison. As shown in Table \ref{tab:action}, with reward modeling and EPO training, our LLM-based interaction module is able to achieve better performance than the hard-coded program. We also observe that without reward modeling, our interaction module fails to achieve comparable result with respect to the deterministic program due to the inaccuracy in choosing low-level actions. We find that since the interaction module in this setup is only trained to imitate previous action trajectories, it fails on the test tasks when the setup is different from the training settings. For example, we would expect the agent to first ``open the drawer" when the drawer is closed before attempting to ``pickup the pen''. However, in training data, the majority of ``pickup object'' actions do not require to open the receptacle object first.

\begin{table}[h]
\small
\centering
\scalebox{0.9}{
\begin{tabular}{ccccc}
\toprule
\ Action Policy & Feedback & Reward & Unseen & Seen \\ 
\midrule
Program & - & - & 0.5383 & 0.4939 \\
\midrule
Model & No & No & 0.2907 & 0.2707 \\
Model & Yes & No & 0.5116 & 0.4744 \\
Model & Yes & Yes & \textbf{0.5542} & \textbf{0.5341} \\
\bottomrule
\end{tabular}
}
\caption{Comparison between static program and learning-based interaction module. Feedback indicates whether we include feedback information. Reward indicates whether we use SFT or EPO with data gathered during interaction.}
\label{tab:action}
\end{table}

\textbf{Environment feedback can help subgoal decomposition.} We use supervised finetuning to fine-tune the subgoal decomposition policy with environment feedback and use the static program as the interaction module as a fair comparison. Both the learning algorithm and the interaction module are the same as the baseline module. As shown in Table \ref{tab:feedback}, with either interaction feedback or visual feedback or a combination of both, we obtain performance gain on both seen and unseen tasks. We also find that a combination of both types of feedback reaches the best performance and that the interaction feedback exhibits more benefit for training than only using visual feedback. One possible reason is that our image captioning model only gives a scene description while the interaction feedback that is more concrete indicator on whether the object is a potential candidate for subgoals. 

\begin{table}[h]
\small
\centering
\scalebox{0.95}{
\begin{tabular}{ccccc}
\toprule
\ Model & interaction & visual & Unseen & Seen \\ 
\midrule
Baseline & No & No & 0.4397 & 0.4036 \\
\midrule
Augmented & Yes & No & 0.5383 & 0.4939 \\
Augmented & No & Yes & 0.4738 & 0.4317 \\
Augmented & Yes & Yes & \textbf{0.5334} & \textbf{0.5036} \\
\bottomrule
\vspace{-0.2in}
\end{tabular}
}

\caption{Comparing different feedback types on validation set. Interaction means whether we include interaction feedback when learning the reward model. Visual means whether we include visual feedback.}
\label{tab:feedback}
\vspace{-0.2in}
\end{table}

\eat{
\begin{figure}[h]
\begin{center}
\centerline{\includegraphics[width=0.45\textwidth]{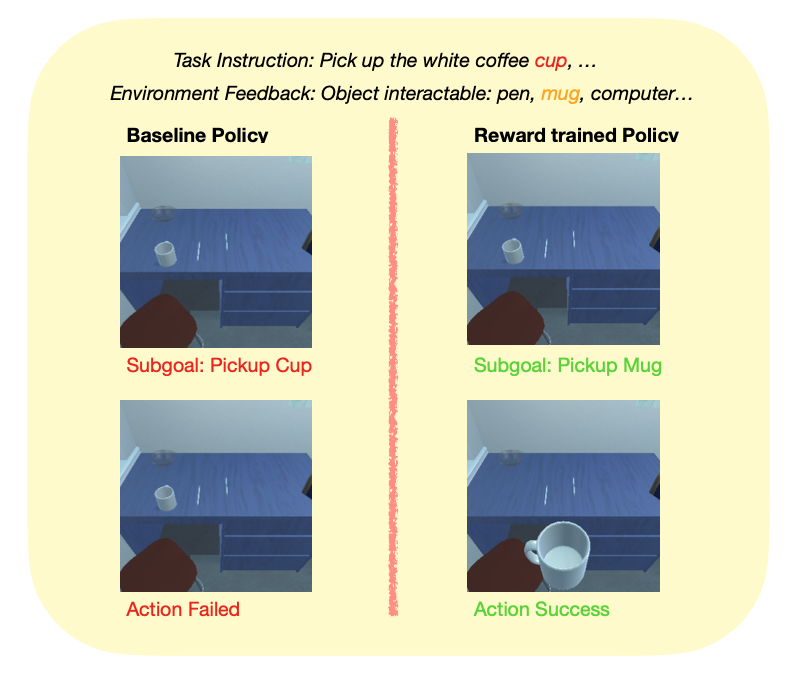}}
\caption{A comparison between the baseline subgoal decomposition policy and EPO's subgoal decomposition module.}
\label{high_policy_vis}
\end{center}
\end{figure}

\begin{figure}[h]
\begin{center}
\centerline{\includegraphics[width=0.45\textwidth]{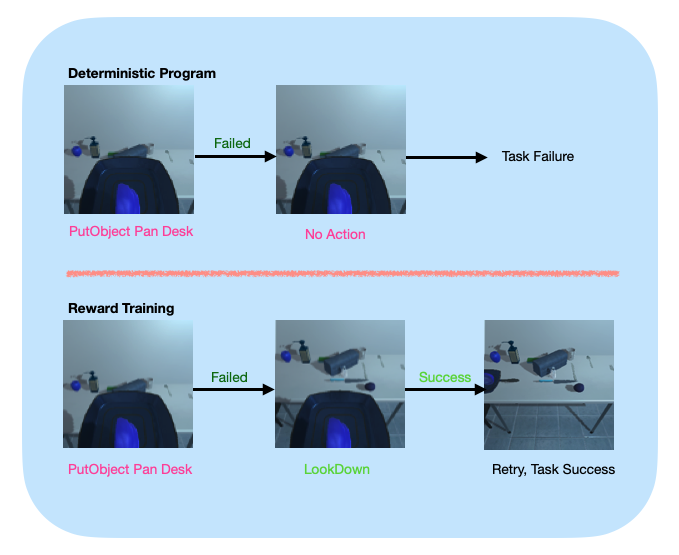}}
\caption{A comparison between the deterministic program and the EPO-tuned interaction module. }
\label{low_policy_vis}
\end{center}
\end{figure}
}

\subsection{Qualitative Analysis}
In addition to quantitative experiments, we visualize the performance of our policies and investigate their effectiveness. Figure \ref{main_vis3}(a) shows a comparison between the baseline policy and the EPO-tuned policy. We see that the baseline policy outputs subgoal predictions closely following the language but outputs the wrong object ``cup'' that the low-level interaction module cannot process. However, from environment feedback we detected ``mug'' exists. Our EPO-tuned policy is able to output the correct parameterization for the subgoal and complete the task. Figure \ref{main_vis3}(b) shows a comparison between the hard-coded deterministic program and our learning-based low-level interaction module. We find that the deterministic program fails because although it outputs the action that is nearly correct but the agent is not close enough to the object so the action (Putobject) cannot be executed. On the other hand, after EPO-tuning our module learn to first output actions to adjust its pose, which leads to success interaction with the environment.

\section{Conclusion}
In this paper, we presented a hierarchical LLM-based framework for long-horizon decision-making tasks, addressing the inherent challenges of extensive planning and the need for scalable training signals. By leveraging a reward model that integrates multimodal environment feedback, and introducing Environment Preference Optimization (EPO), we successfully generated training signals for unannotated datasets. Our framework demonstrated state-of-the-art performance on the ALFRED benchmark. Future work will focus on exploring the integration of additional types of multimodal feedback to further enhance the agent's decision-making capabilities, as well as extending our framework to real world robotics tasks. 
\clearpage
\section*{Limitations}
We evaluate the proposed method on ALFRED, where the low-level action space is discrete and annotated with language. For some continuous control tasks, the action space can be much larger and hard to interpret. Future work will focus on exploring the integration of additional types of multimodal feedback to further enhance the agent's decision-making capabilities, as well as extending our framework to real world robotics tasks. 

\section*{Acknowledgement}
This work was conducted using computational resources and services at the Center for Computation and Visualization, Brown University. The project was in part supported by ONR grant \#N00014-22-1-2592, and the Samsung Global Research Outreach program. The authors would like to thank Calvin Luo, Tian Yun, as well as the anonymous reviewers for valuable feedbacks.

\bibliography{main}

\clearpage

\appendix

\section{Symbolic Representation and Prompt Examples}
\begin{figure}[h]
\begin{center}
\centerline{\includegraphics[width=0.5\textwidth]{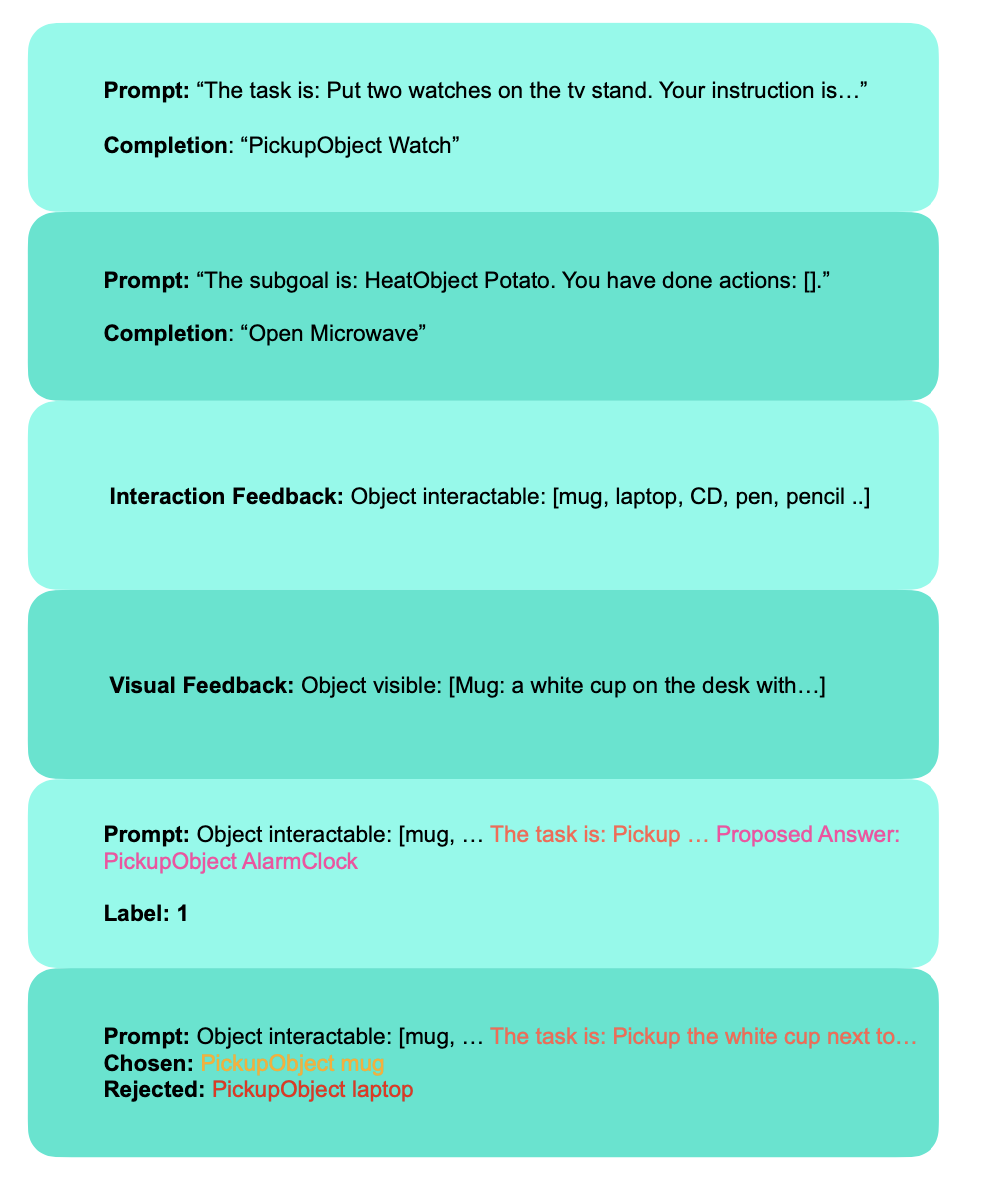}}
\caption{A illustration of prompt to our LLM policies. From top to bottom: example of baseline subgoal policy, example of baseline interaction policy, example of interaction feedback , example of visual feedback , example of reward model training Data, example of Environment Preference Data
}
\label{prompt}
\end{center}
\end{figure}

In dealing with multimodal feedback information, it is crucial for us to design structure prompt to interact the LLMs. Luckily, the task specifications $\tau$, subgoal and low-level action annotations are already in the form of text so we do not need to further tune them. The visual and interaction feedback however, needs to proper symbolically represented. For example, when our object detector finds visible objects, our agent will interact with it. If the attempted interaction is successful, our agent will receive a boolean value from the system. We would describe this event as ``action successful'' for our low-level policies. In gathering the environment feedback, we would just simply append the name of the object to the existing object list. Visual feedback, which is the image captioning data, is already in the form of text. Figure \ref{prompt} illustrates prompt examples of our pipeline. 
\begin{algorithm}
\small
   \caption{Environment Preference Dataset Generation}
   \label{alg:epo_data}
\begin{algorithmic}[1]
   \STATE {\bfseries Input:} Task specification $\tau$, Environment Feedack specification $f$, Possible Outputs $P$ 
   \STATE // \textbf{Initialization} \\
   Initialize environment preference ranking as empty list. \\
    Initialize environment preference dataset as empty list. \\
   \FOR{$p_i$ $\gets$ possible outputs $P$}
   \STATE $r_i$ $\gets$ Reward Model($f$,$\tau$,$p_i$)
   \STATE Append $p_i : r_i$ to environment preference ranking
   \ENDFOR
    \STATE Sort environment preference ranking according to $r_i$\\
    \FOR{$p_i$ $\gets$ environment preference ranking[1:]}
    \STATE preference data point $\gets$ \{'prompt': $f$, $\tau$, 'chosen': $p_0$, 'rejected': $p_i$\}
   \STATE Append preference data point to environment preference dataset
   \ENDFOR
\end{algorithmic}
\end{algorithm}

\section{Additional Algorithm Details}
\label{app:imp}
In Algorithm~\ref{alg:epo_data}, we provide the detailed steps of our environment preference data generation process. We first infer reward values from possible outputs from the policy using the reward model. Then we rank all the possible outputs based on reward. Then we pick the output with the highest reward as the chosen prompt and the rest as the rejected output, the prompt is environment feedback $f$ prepend to task specification $\tau$. 

\textbf{Language Model Training} For all our policies, we use pretrained Llama-7B as the backbone LLM. It has around 7 billion parameters. All our experiments are conducted on NVIDIA A6000 GPU. We use LoRA to efficiently fine-tune the language models with the datasets we design. Specifically, we use $r = 8$, $\alpha = 32$, and lora dropout equals to 32. We use a learning rate of $1e-5$ and adam optimizer. The target fine-tuning modules are the q-projection layers and the v-projection layers. Approximately, 5\% of the parameters are trained. We find our training usually coverges within 1 epoch. For EPO training, the learning rate is set to $1e-6$. In all our training, we use a batch size of 32. To train the reward model, we use Llama2 with a classification head instead of casual generation. For BLIP-2, we only use the image as input to generate the captions. We did try providing additional text in the prompt but did not observe any clear benefits to the results.

\textbf{ALFRED} There are two categories of subgoals, navigation and interaction. We use the deterministic navigator provided by \citep{shridhar2020alfworld}, which needs the view-point location to navigate to. However, we did not use environment meta information to obtain the view-points for the objects. Our agent exploration process is able to successful record possible view-points for successful interaction. The only meta information we use is action success and agent inventory. To determine the object to navigate to, we use the target object of the next subgoal as the navigation target. To interact with objects in ALFRED, one needs to output a interaction mask. We do so using the MaskRCNN model provided by \citep{pashevich2021episodic}. We use the checkpoints from Episodic Transformer \citep{pashevich2021episodic}.

\textbf{Environment exploration} In order to receive feedback from the environment, we need an structured process of exploration. First, we define the concept of ``view-points", which indicates the location, direction and camera angle. A view-point is parameterized with four variables $x, y, r, h$. $x$ and $y$ indicates the grid coordinates of the agent in the 3D-environment. $r$ indicates the direction which our agent is facing. $h$ indicates the eye level angle of our agents. We consider the height of our agent fixed at all time. We explore the environment to let the agent visit as much view-points as possible. We allow agents to explore all possible locations and ``view-points'' to interact with the visible objects. Through our exploration, we apply object detector to obtain the visible objects. We record the object that our agent successfully interacted with. After exploration, we will have a ``view-point'' point map of all objects the agent has interacted with.


   

\textbf{Decomposition module} The input of our decomposition module is the task instructions and the output is generated text that indicates the subgoal prediction. The generated text will be post-processed into high-level actions and target objects in the form of texts. One could form this problem as a classification task without the intermediate text. But we argue that generating free-form language generalizes better to environments and tasks when the possible subgoals of our agent are hard to be defined in a closed set.

\textbf{Interaction module.} After our agent predicts the subgoals, it uses an interaction module to output the low-level actions to complete each subgoal sequentially. There exists two types of subgoals: navigation and interaction. For a navigation subgoal, we use a view-point-based navigation planner with the object location information we gained during agent exploration. For interaction subgoals, as noticed by previous work \citep{min2021film}, the action sequences required to complete them can be quite deterministic and is possible to solved them with a static program. Nevertheless, we propose a learning-based method in which our model uses a large language model as its backbone. It takes in the subgoal information, the interaction feedback from its previous action, and its historical actions in completing this subgoal, all symbolically-represented in text and outputs the next low-level action. Our model generalize better to the scenarios which action dynamics are less deterministic. It predicts the next action based on interaction feedback and previous actions in an auto-regressive manner. Later in experiments, we show that this learning-based module can be further improved with environment feedback and EPO.

\textbf{Reward Modeling} Recall that our reward model estimates the likelihood of the output is correct and form the environment preference dataset through ranking. In training the reward model for the subgoal decomposition module, we use the annotated dataset to form input consist of environment feedback $F$, task specification $T$, and proposed answer $P$. Then we label the correct annotation with 1 to form a positive pair and randomly select an incorrect output from each of the parameters to form 2 negative pairs. After we train the reward model, we make inference on the unannotated dataset where $F$ and $T$ are available but the proposed answer is from our SFT pretrained module or other possible outputs. Then we can form the preference dataset by comparing the reward between proposed answers given the same $F$ and $T$. In training the reward model for interaction module, we gather online data by allowing our agent to attempt various pose changes and interactions until it could succeed its intended action. Then we record the actions led to successful interaction and other unsuccessful actions to form the positive and negative pairs. Then the process to form the preference dataset is similar with that of the subgoal decomposition module. We did not any AI assistant in writing this paper.

\textbf{Baselines}
We compare the overall performance of our framework with the state-of-the-art methods on ALFRED public leaderboard. We obtain all the results following the standard setting in ALFRED where we first let the agent learn from the given dataset offline, and then test the online rollout performance of the learned policies (modules) on the given set of new test tasks. All baselines have access to the same amount of information, as this is the standard setting required by ALFRED to get a score on the public leaderboard. Thus we believe the comparison with all the baselines is fair. We will add more descriptions for each baseline listed in the updated version of our paper as suggested. Specifically, HLSM proposes to build a persistent spatial semantic representation from natural language instructions. FILM involves the creation of a semantic map of the environment and a semantic search policy to navigate and interact based on the instructions provided. EPA uses a discrete graph representation enriched with new perceptions during exploration, allowing the agent to generate new planning problems and recover from action failures. Prompter introduces a method that replaces the traditional semantic search module in embodied instruction following systems with language model prompting. CAPEAM enhances an agent's ability to perform household tasks by integrating semantic context and maintaining the state of objects within the environment.
\begin{table}[h]
\small
\centering
\scalebox{1.0}{
\begin{tabular}{cc}
\toprule
Methods & Success rates \\ 
\midrule
Baseline (without environment feedback) & 0.7409 \\
EPO (with 10$\%$ annotated data) & 0.9781 \\
EPO (with fully annotated data) & 0.9905 \\
\bottomrule
\end{tabular}
}
\caption{Results on BabyAI}
\label{tab:babyai}
\end{table}

\textbf{Results on BabyAI}
We also conduct a set of experiments on BabyAI~\citep{DBLP:conf/iclr/Chevalier-Boisvert19} minibosslevel, which is an environment where an agent navigates and interacts in a grid world to achieve a goal described in language. As shown in Table~\ref{tab:babyai}, we observe that EPO with environment feedback (object type observed by the agent) can boost task success rate from 0.7409 to 0.9905 and with 10\% of labeled data and EPO, our policy can reach 0.9781 task success rate, which is just 0.0124 less than using all labeled training data.

\end{document}